\documentclass[10pt,twocolumn,letterpaper]{article}

\usepackage{wacv}              

\usepackage{graphicx}
\usepackage{amsmath}
\usepackage{amssymb}
\usepackage{booktabs}

\usepackage[pagebackref,breaklinks,colorlinks]{hyperref}

\usepackage[capitalize]{cleveref}
\crefname{section}{Sec.}{Secs.}
\Crefname{section}{Section}{Sections}
\Crefname{table}{Table}{Tables}
\crefname{table}{Tab.}{Tabs.}

\usepackage{times}
\usepackage{epsfig}
\usepackage{graphicx}
\usepackage{amsmath}
\usepackage{amssymb}

\usepackage[T1]{fontenc} 

\begin{document}

\title{VDD: Varied Drone Dataset for Semantic Segmentation}

\author{Wenxiao Cai \hspace{0.5em} Ke Jin  \hspace{0.5em} Jinyan Hou  \hspace{0.5em} Cong Guo  \hspace{0.5em} Letian Wu  \hspace{0.5em} Wankou Yang\thanks{Corresponding author: wkyang@seu.edu.cn} \\
Southeast University
}

\maketitle

\begin{abstract}
Semantic segmentation of drone images is critical for various aerial vision tasks as it provides essential semantic details to understand scenes on the ground. Ensuring high accuracy of semantic segmentation models for drones requires access to diverse, large-scale, and high-resolution datasets, which are often scarce in the field of aerial image processing.
While existing datasets typically focus on urban scenes and are relatively small, our Varied Drone Dataset (VDD) addresses these limitations by offering a large-scale, densely labeled collection of 400 high-resolution images spanning 7 classes. This dataset features various scenes in urban, industrial, rural, and natural areas, captured from different camera angles and under diverse lighting conditions.
We also make new annotations to UDD\cite{udd} and UAVid\cite{uavid}, integrating them under VDD annotation standards, to create the Integrated Drone Dataset (IDD).
We train seven state-of-the-art models on drone datasets as baselines.
It's expected that our dataset will generate considerable interest in drone image segmentation and serve as a foundation for other drone vision tasks. Datasets are publicly available on our website at \href{here}{https://github.com/RussRobin/VDD}.
\end{abstract}


\section{Introduction}
Visual scene understanding of drone images has sparked great interest in the computer vision community, as it presents new challenges and potential for high-resolution drone image processing and complicated drone vision tasks like depth estimation\cite{uaviddepth21,depth1,depth3}, image stitching~\cite{ISEO,obj-gsp,stitch1,stitch2}, 3D reconstruction\cite{3destimation1,3destimation2}, obstacle avoidance\cite{obstacleavoid1} and adaptive path planning\cite{pathplan1,pathplan2}. 
We propose Varied Drone Dataset(VDD), a varied and large dataset of 400 high-resolution images to facilitate future research in semantic segmentation of drone images, and we believe that VDD will pave the way for new approaches in the field of aerial image processing.
\begin{figure}[h]
	\begin{center}
    \includegraphics[width=1\linewidth]{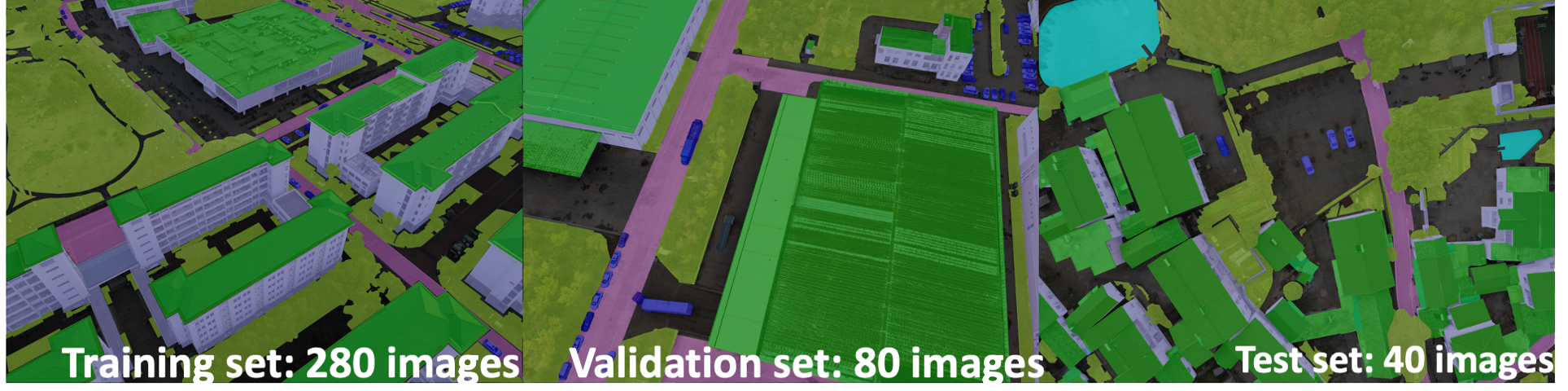}
	\end{center}
 	\caption{Sample images in VDD train/val/test set. The three images provide a glimpse into the variance in VDD: they are taken in urban, rural and industrial areas respectively, and the camera angles are 30, 60 and 90 degrees.}
	\label{fig_head}
 
\end{figure}

Semantic segmentation in autonomous driving and remote sensing has always been a heated topic, and many large datasets have been released to better understand objects on the ground from the perspective of cars\cite{cityscapes}, satellites or remote sensors \cite{isprs}. However, there are few datasets that focus on semantic segmentation of low-altitude drone images. Existing drone datasets are small, and mainly focus on urban scenes. To further help drones understand all kinds of scenes with varying conditions, we propose VDD to enrich  drone datasets. VDD focuses on the diversity of images, encompassing not only urban scenes but also a wide range of other scenarios. Additionally, VDD incorporates variations in camera angles, weather conditions, and lighting situations.
Fig. \ref{fig_head} contains the example images of training/validation/test set of VDD. With domain gap, dataset quality and copyright matters in consideration, we also make new annotations to UDD\cite{udd} and UAVid\cite{uavid} and fuse them under VDD annotation standards. This Integrated Drone Dataset (IDD) contains 811 images in 7 semantic classes, being the largest and most comprehensive drone dataset for semantic segmentation, and is twice the size of VDD.

\begin{figure*}[h]
	\begin{center}
		\includegraphics[width=1\linewidth]{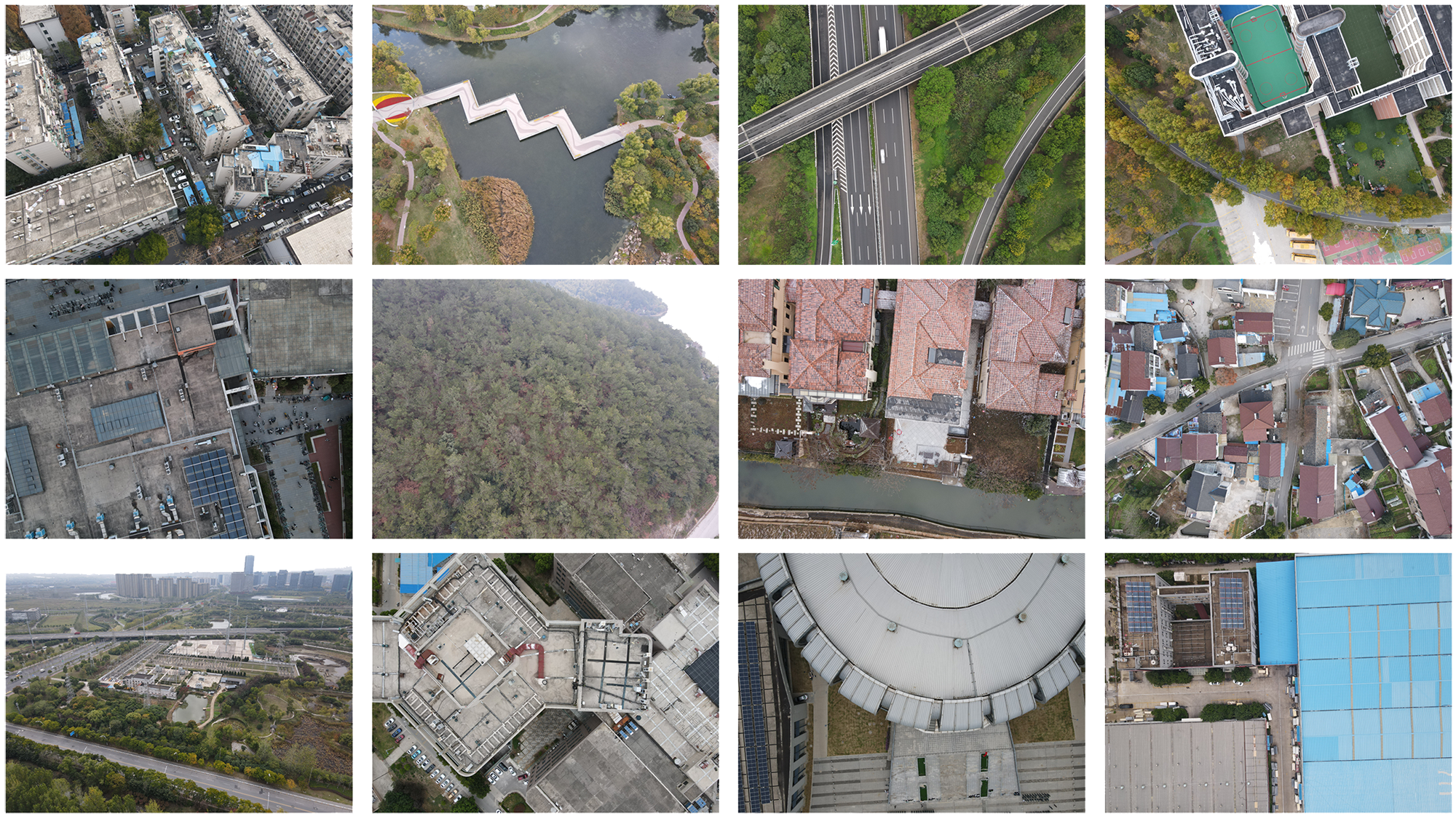}
	\end{center}
	\caption{Various scenes in VDD. From left to right, top to down: urban residence, lake, highway, highschool, canteen in university, mountains, villa zones, rural villages, transformer substation, hospital, gym and factory.}
	\label{fig_scene}
\end{figure*}

To summarize, the main contributions of the new dataset include:
\begin{itemize}
    \item A new Varied Drone Dataset(VDD) of 400 high resolution images in seven semantic classes, featuring variation in images.
    \item New annotations to two existing datasets to fuse them with VDD. The Integrated Drone Dataset(IDD) is so far the largest public available drone dataset.
    \item We train state-of-the-art models to set new baselines on drone image segmentation.
\end{itemize}

\section{Overview of Drone Datasets}
Large-scale and diverse datasets have played a crucial role in recent advances in the computer vision community.
Cityscapes~\cite{cityscapes}, which labels 25000 images at 1024x2048 pixels, is useful for understanding street scenes and has applications in autonomous driving. 5000 images in Cityscapes~\cite{cityscapes} are provided with fine labels in 30 classes, and 20000 images are with 8 categories of coarse annotation. In the remote sensing community, numerous semantic segmentation datasets exist, such as the ISPRS 2D semantic labeling dataset~\cite{Rottensteiner2014ResultsOT}, which provide aerial images of cities at either 6000x6000 pixels or 2000x2000 pixels, with resolutions of 5cm or 9cm. 
These large datasets focus on driving and remote sensing make it possible to conduct studies in visual scene understanding. Such fundamental researches make high-level applications like autonomous driving and automatic exploration of land resources possible.

However, there are only a few semantic segmentation datasets available for drone images captured at low altitudes (no higher than 120 meters). 
To the best of our knowledge, there are four datasets, including Aeroscapes~\cite{aeroscapes}, ICG Drone Dataset~\cite{icg}, UAVid~\cite{uavid}, and UDD~\cite{udd}. 
Aeroscapes provides images of 720p resolution, while the remaining three datasets offer 4k or higher resolution images, but their sizes are small. FloodNet~\cite{floodnet} contains aerial images of 4000x3000 pixels and focuses on disaster assessment, primarily on the effects of floods on homes. As discussed in Cityscapes~\cite{cityscapes}, special conditions such as natural disasters require specialized methods, so FloodNet is out of the scope of the paper.

\begin{figure}[h]
	\begin{center}
		\includegraphics[width=1\linewidth]{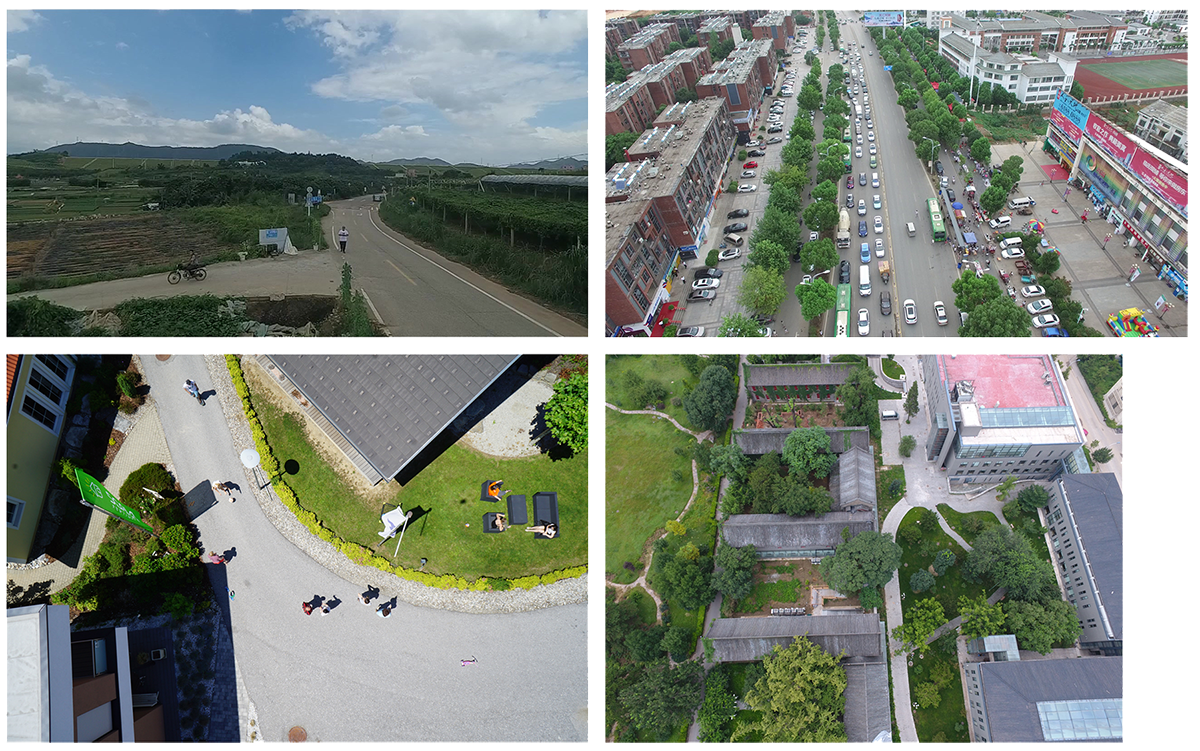}
	\end{center}
	\caption{Typical scenes in Aeroscapes, ICG Drone Dataset, UAVid and UDD. }
	\label{fig_otherdata}
\end{figure}

Aeroscapes \cite{aeroscapes} dataset was inspired by Cityscapes and assumes that drones can provide more information than cameras on cars. However, the scenes captured in Aeroscapes are often repetitive and lack satisfactory resolution. 
ICD Drone Dataset \cite{icg} is designed to assist 3D reconstruction of single buildings, so the drones fly low. Scenes in it are relatively simple.
The complexity of a scene is related to factors such as image size, resolution, number of objects in an image, length of boundaries between different categories, and variation of sizes and shapes within a class. 
After reviewing the images in these four drone datasets, it can be qualitatively determined that UAVid \cite{uavid} and UDD \cite{udd} have more complex scenes than ICG Drone Dataset \cite{icg} and Aeroscapes 
\cite{aeroscapes}. 
UAVid \cite{uavid} labels 300 images in eight categories with 4k maps and is firther used by its team to perform self-supervised learning of depth maps from video, which can be applied in 3D reconstruction~\cite{uaviddepth21}. 
UDD \cite{udd} is specifically designed to assist 3D reconstruction.
UDD \cite{udd} thus distinguishes between roof and walls to help with an improved Structure From Motion (SFM)\cite{sfm} method. 
It is important to note that existing drone datasets focus on urban scenes, while agricultural and industrial zones, mountains, and water are neglected. In contrast, remote sensing datasets like Loveda, GID, and MiniFrance~\cite{loveda,gid,minifrance} focus on land resource identification and utilization but lack the resolution and detail provided by low-altitude drones. To help drones understand various unmodeled conditions, we aim to create a diverse dataset with varied camera angles, scenes, and light conditions captured at altitudes ranging from 50 to 120 meters. Fig.~\ref{fig_scene} shows some of scenes included in VDD, while Fig.~\ref{fig_otherdata} displays images from the other four drone datasets.

\section{The Proposed VDD Dataset}
\begin{figure}[t]
	\begin{center}
		\includegraphics[width=1\linewidth]{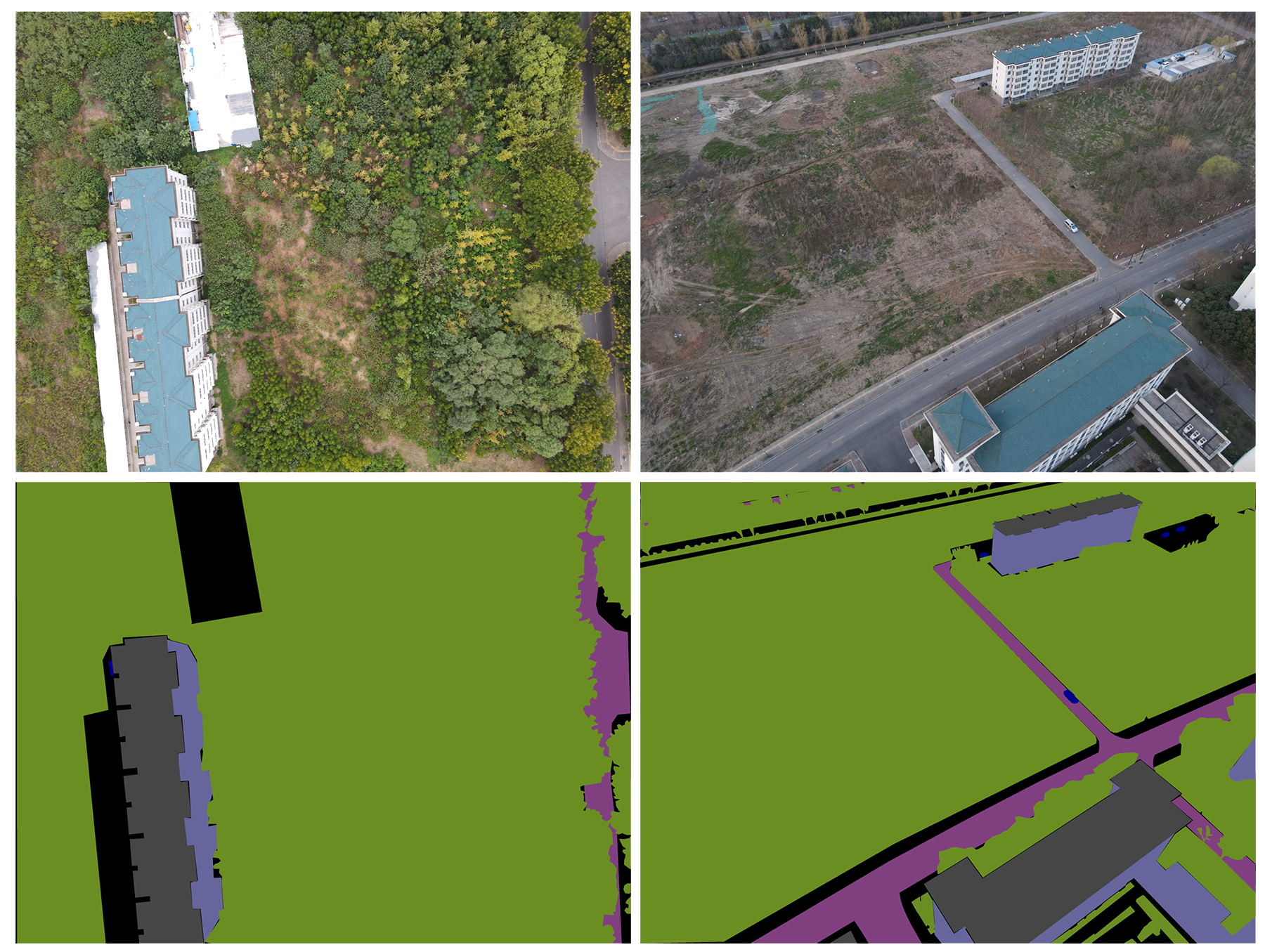}
	\end{center}
	\caption{These images were taken in spring and autumn, respectively. The light conditions and vegetation ratios are changed, while the building looks the same.}
	\label{fig_time}
\end{figure}

\textbf{Dataset Collection:} We used DJI MAVIC AIR II for the collection of our VDD dataset: 400 3-channel RGB images with 4000*3000 pixel size. 
We went to 23 locations in Nanjing (east China) to collect images: downtown of Nanjing(urban residential and commercial area), Jiangning district (urban-rural fringe area, industrial zones and natural landscapes), and Lishui district (rural area, industrial zones, large transportation hubs and natural landscapes). 
For a hospital, we conducted two group of shoots at the same location, one in the morning and one in the evening. For a villa residential area and a university campus, we conducted shoots during different seasons. Fig. \ref{fig_time} shows scene variance of a campus building in spring and autumn. As a result, our dataset encompasses diverse scenes, lighting conditions, seasons, and times. We discuss in detail how varied VDD is in Section 3.1 .

All the images were taken with drones flying at altitudes ranging from 50m to 120m. The high-resolution images captured at this altitude ensure that both abundant details and scene complexity are present. Compared to the ICG Drone Dataset \cite{icg}, each image in VDD contains a much greater variety of scenes, objects, and complexity of boundaries, while also capturing clear details on the ground. 

\textbf{Pixel-level Labeling:} LabelMe\cite{labelme} is used as the annotation tool to obtain 400 densely labeled images. Labeling a large number of pixels can be both time-consuming and labor-intensive. Most images in VDD took about 3 hours to label. In Integrated Drone Dataset (IDD), we label water in 141 UDD\cite{udd} images, and roof and water in  270 UAVid\cite{uavid} images. This takes about 0.5 hour per image. Our VDD dataset was labeled by a team of engineers. We randomly selected 20 images to label repeatedly by all engineers. Our average pixel overlap rate is higher than 99\%, indicating a high annotation accuracy. 

Recently, interactive segmentation \cite{FocalClick, FirstClick, cfr-icl, eiseg} has gained significant popularity, and the Segment Anything Model \cite{sam} has also garnered considerable attention. During the annotation process, we attempted interactive segmentation; however, we observed suboptimal results. 

\subsection{VARIED dataset}
The diversity of a dataset is crucial for successful network training. We consider 3 important variations in our VDD dataset.

\begin{figure*}[h]
	\begin{center}
		\includegraphics[width=1\linewidth]{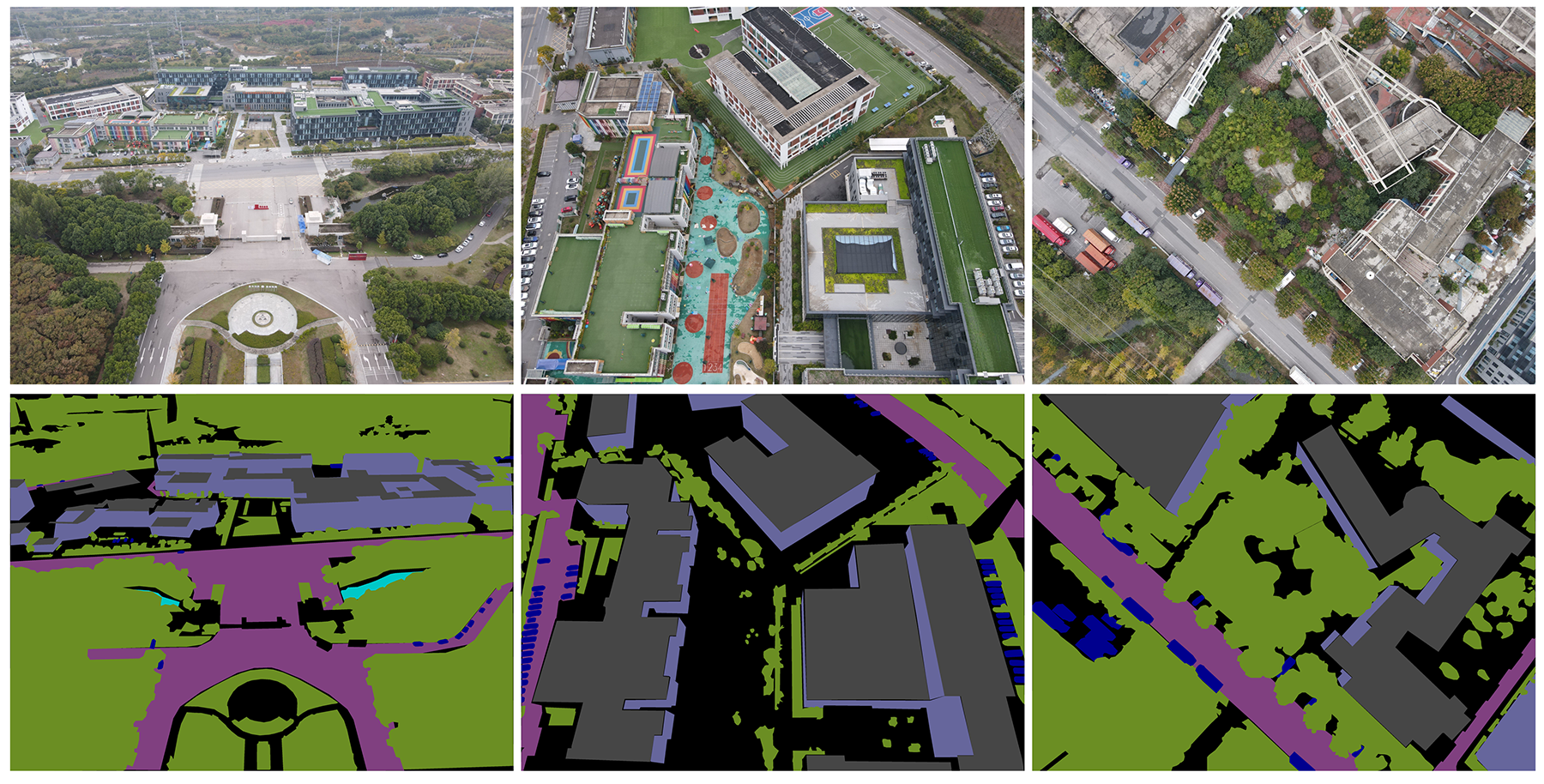}
	\end{center}
	\caption{The three images are taken with three camera angles at the same place, including 30, 60 and 90 (bird view) degrees.}
	\label{fig_angle2}
\end{figure*}

\paragraph{Varied camera angles.} We consider this variation as a highly effective method of data augmentation. While traditional data augmentation techniques such as flipping, rotating, cropping, scaling, panning, and dithering can modify the appearance of an object, they cannot replicate the effects of changes in camera angles. With varying camera angles, each object undergoes geometric transformation between images, while retaining its color and texture. This results in a more diverse and realistic dataset. Additionally, changes in shading relationships between objects occur with changes in camera angles. To achieve maximum variation in our dataset, VDD was created using camera angles set at 30, 60, and 90 degrees. 90 degrees is a bird view, watching vertically from air to ground. The group of images shown in Fig. \ref{fig_angle2} aim at the same buildings, but with different camera angles. Thus they contain significantly various semantic information, such as variation of scales, and difference in occlusion relationships. In detail, there are 69 images taken at 30 degrees, 75 taken at 60 degrees, and 256 taken at bird view.

\begin{figure}[t]
	\begin{center}
		\includegraphics[width=1\linewidth]{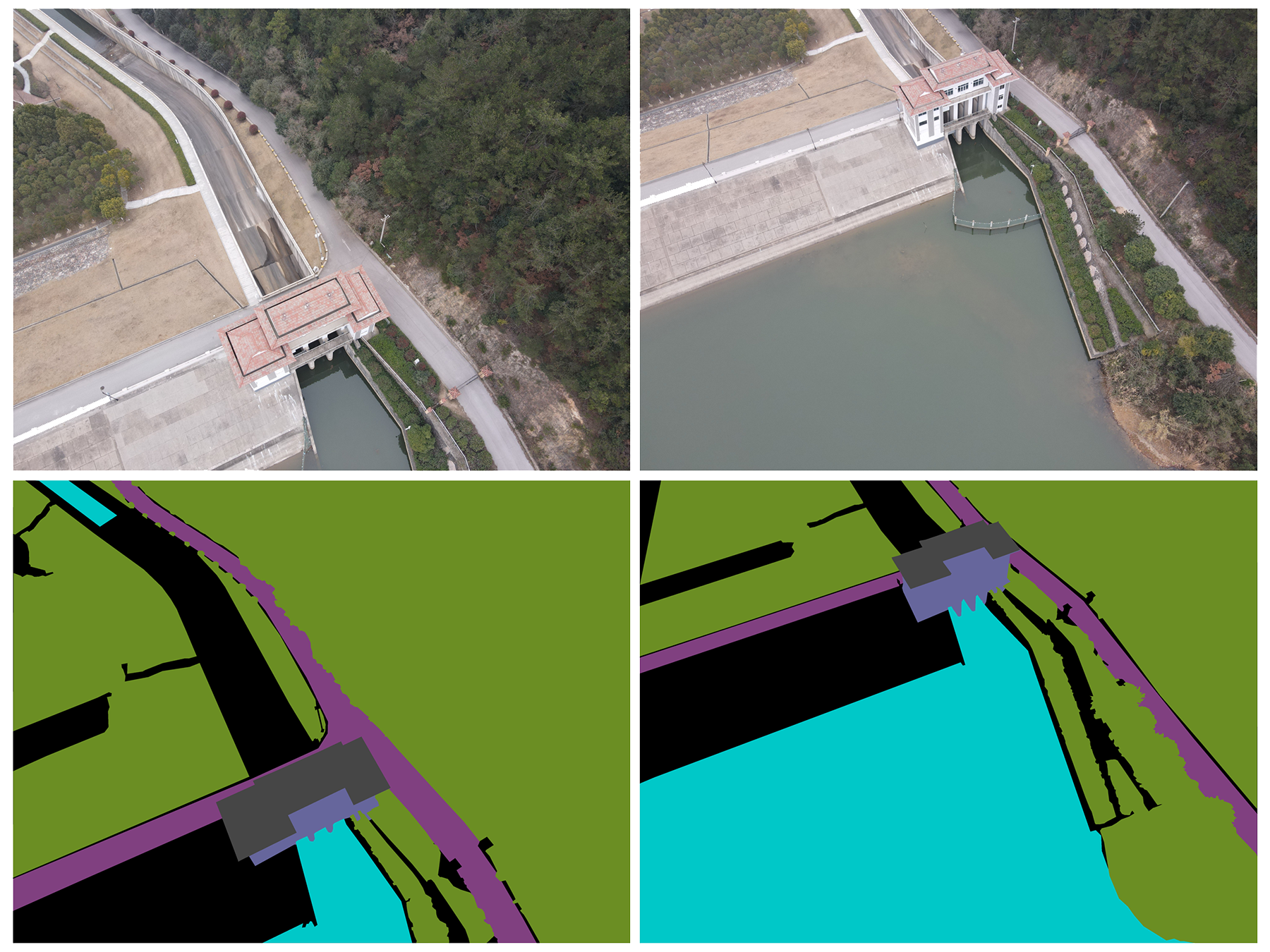}
	\end{center}
	\caption{Drone provides different views of dam, building and lake.}
	\label{fig_angle1}
\end{figure}

\paragraph{Varied scenes.} Our dataset is large, not only in the total pixels, but also in its scene complexity. Since other datasets \cite{uavid,udd,icg,aeroscapes} have already collected many urban scene, we pay more attention on industrial factories, rural areas and natural landscapes, while also including urban scenes.
We aim to cover as many semantic environments as possible in this dataset. Specifically, VDD contains the following scenes:
\begin{itemize}
\item Municipal residential zones: 37 images
\item Urban facilities like libraries, hospitals: 63 images
\item A tall hospital with parking lots, shot during sunrise and sunset: 32 images
\item Schools and college buildings: 69 images
\item Highways, roads and parking lots: 24 images
\item Low-height urban facilities like gyms and canteens: 53 images
\item Industries and factories: 31 images
\item Natural scenes: rivers, lakes and mountains: 51 images
\item Rural areas: villages, farm fields and rural kindergartens: 23 images
\item Rural villa zones: 17 images
\end{itemize}
Fig. \ref{fig_scene} demonstrates 12 scenes in VDD, taken in 12 locations. Note that it's not rigorous to classify one image into a single category, since scenes are complex. For example, roads, parking lots are both in schools, buildings and villas are sometimes around lakes... 
The above image count are based on where we took photos. For example, every image taken around a factory is categorized into industries, although roads and farmlands are just beside the factories. 
To make our dataset as varied as possible, we include different standpoints of view in VDD. When a drone flies over a building, it can capture all aspects of that object, as shown in Fig.~\ref{fig_angle1}. Following previous works \cite{uavid,udd,floodnet,aeroscapes,icg}, our drone doesn't follow fixed routes whenn flying. We simply shooting at different locations within the selected area.

\paragraph{Varied weather and light conditions.} We shot at various times of the day, from early morning to late afternoon. Images are collected in daytime from Summer 2022 to Spring 2023, while nights are not considered in VDD.  The pair of images in Fig. \ref{fig_time} were taken in nearby areas but in different seasons. Vegetation and some ground views changes in the dataset. 
For a villa zone, we took photos both in summer and winter. For a hospital, we shoot in different times fo a day. This is to provide variations in weather and light conditions.
All weather conditions with acceptable light condition to see the ground clearly are allowed in VDD. Extreme weather where drones can't take off is not included, like snowy and rainy days.

\subsection{Volume of Dataset}
\begin{table*}[h]
	\begin{center}
		\begin{tabular}{|l|c|c|c|c|c|}
			\hline
			Dataset & Drone altitude & Image size & Number of classes&Volume & TPV \\
			\hline\hline
			Cityscapes (fine) & Taken on cars & 1024x2048Px & 30 & 5000 & 10.5B \\
   			Cityscapes (coarse) & Taken on cars & 1024x2048Px & 8 & 20000 & 42B \\
			Aeroscapes & 5 to 50m & 1280x720Px & 12 & 3269 & 3.0B \\
			ICG Drone Dataset & 5 to 30m & 6000x4000Px & 20 & 600 & 14.4B\\
			UAVid & 50m & 4096x2160 or 3840x2160Px & 8 & 300 & 2.6B\\
			UDD & not given	& 3000x4000 or 3840x2160Px & 6 & 141 &  1.5B\\
			FloodNet & not given & 3000x4000Px & 10 & 2343 & 28.1B \\
			\textbf{VDD (ours)} & 50 to 120m & 3000x4000Px & 7 & 400 & 4.8B\\
			\textbf{IDD (ours)} & varies & varies & 7 & 811 & 8.2B \\
			\hline
		\end{tabular}
	\end{center}
	\caption{Total Pixel Volume (TPV), image size and drone altitude in 7 datasets. B stands for billion, and Px means Pixels in size. Fine annotation of Cityscapes is considered here. Integrated Dataset is the combination of VDD, UDD and UAVid, under VDD's segmentation standards. }
  	\label{table_volume}
\end{table*}

\begin{figure}[h]
	\begin{center}
		\includegraphics[width=1\linewidth]{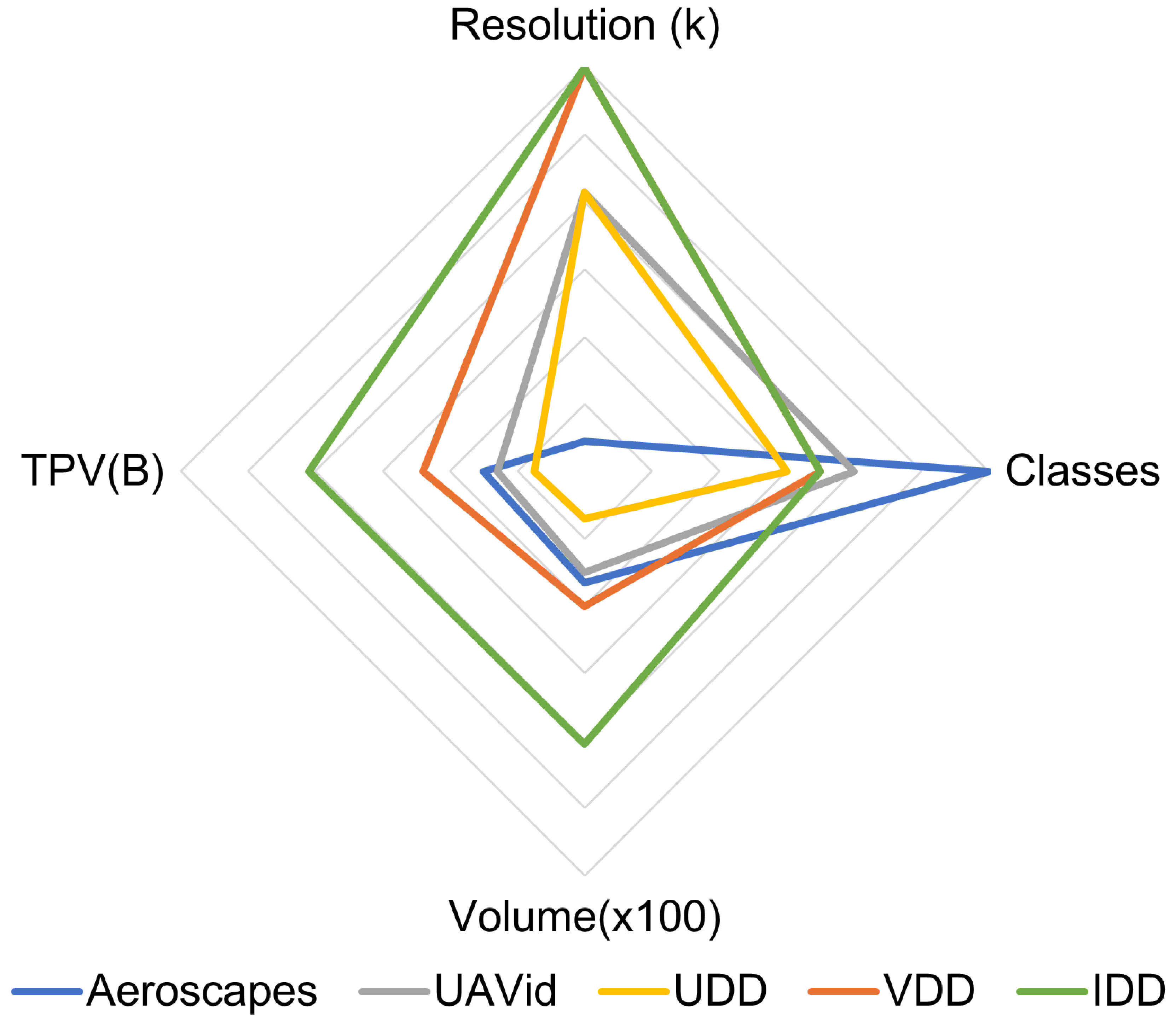}
	\end{center}
	\caption{Comparasions of low-altitude drone image segmentation datasets, where drones fly 50 to 120 meters. We compare image resolution, total pixel volume (TPV), dataset volume and number of classes.}
	\label{fig_datasets}
\end{figure}

Considering number of images and total pixels as shown in Fig.~\ref{fig_datasets}, our proposed dataset is comparable to all existing drone datasets. Table \ref{table_volume} counts total pixel volume (TPV) in Cityscapes\cite{cityscapes} and existing drone datasets. The Cityscapes fine annotation consists of 30 classes and contains 5,000 images. This large-scale dataset was captured from a vehicle perspective. Additionally, Cityscapes also provides 20000 images with coarse annotation. These images are weakly annotated and are typically not directly used for supervised learning. Instead, they require special methods for processing and utilization\cite{coarse}.
When it comes to datasets captured by drones, both UDD \cite{udd} and UAVid \cite{uavid} do not match the size of our proposed VDD. This holds true in terms of both the number of images and the total pixel volume (TPV) contained in the dataset.
Although the ICG Drone Dataset is twice larger as VDD, the information in it is not comparable to ours, because thier dataset mainly contains relatively simple scenes. Our drone flies higher, and the captured scenes are far more complex. Integrated Drone Dataset (IDD) of VDD, UDD\cite{udd} and UAVid\cite{uavid} comprises 811 high-resolution images. 

\begin{figure}[h]
	\begin{center}
		\includegraphics[width=1\linewidth]{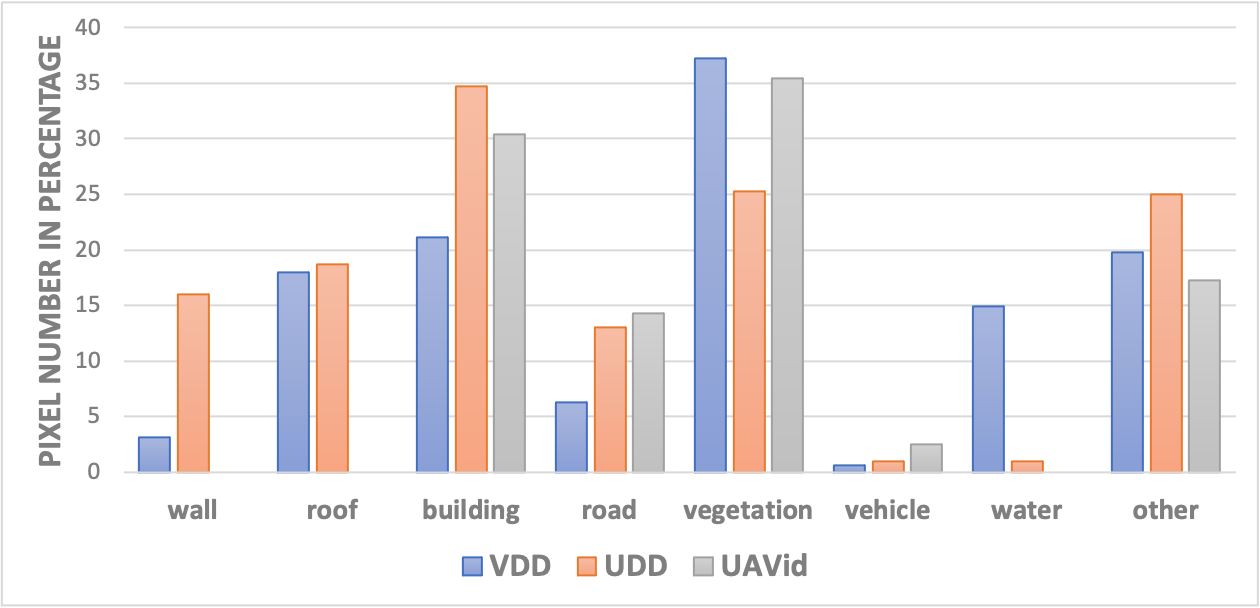}
	\end{center}
	\caption{Number of pixels (in percentage) of 3 drone datasets: VDD, UDD \cite{udd} and UAVid \cite{uavid}. Buildings refer to the sum of roof and wall. Since roofs, walls and water are not categorized in UAVid, we leave the data blank.  }
	\label{fig_pixel}
\end{figure}
\subsection{Class definition}
\begin{figure}[h]
	\begin{center}
		\includegraphics[width=1\linewidth]{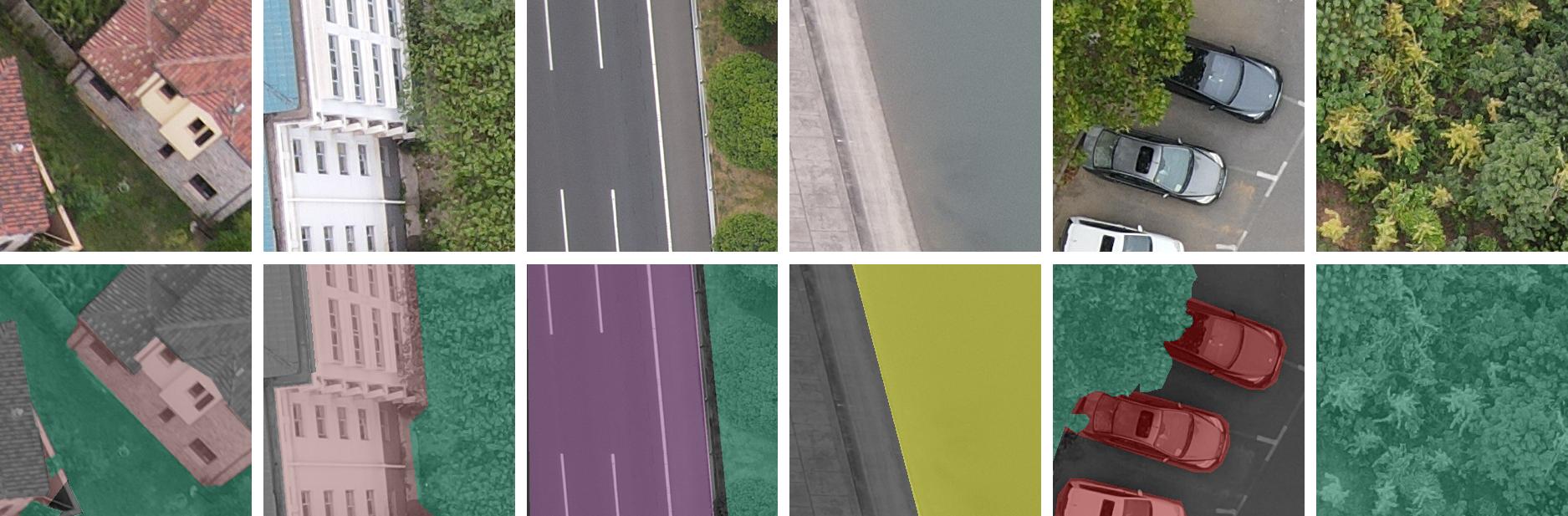}
	\end{center}
	\caption{Class definition in the VDD dataset. From left to right: roof, wall, road, water, vehicle, and vegetation. Note that we label images in pixel-level. This figure is a closs look of annotations in VDD.}
	\label{fig_6class}
\end{figure}

All the pixels of an image in VDD are categorized into 7 classes: Wall, Roof, Road, Water, Vehicle, Vegetation and Others, shown in Fig. \ref{fig_6class}. Labelme \cite{labelme} is used as the annotation tool. The labeling protocol is as follows:
\begin{enumerate}
\item Roof and Wall: Top roof of a building is categorized into Roo. The remaining exterior surfaces of buildings are considered Wall. The open space in the middle of the courtyard is labeled as Walls too. 
The aim is to make our labeling protocol similar to that of UDD.
\item Vehicle: Motor vehicles only. Cars, vans are marked as Vehicle. We label bicycles as Others.
\item Road: Only the roads where vehicles can and are legally allowed to drive on are counted as Road. Some small ways where only people can walk on are excluded.
\item Vegetation: Trees and lower vegetation are both labeled as Vegetation. Soil surfaces are not vegetation, but small soil areas surrounded by grass are labeled as Vegetation too.
\item Water: Surfaces of water, which includes lakes, rivers, pools, and puddles on roads or farmlands.
\item Others: Pixels not belonging to the above 6 classes are labeled as Others, such as humans, playgrounds, bridges and bicycles.
\end{enumerate}
VDD has a wide range of images of cities, factories, rural areas and natures. Thus vegetation and water domains a great part in VDD. UDD and UAVid mainly focus on urban areas. Fig. \ref{fig_pixel} demostrates pixels comparasion (in percentage) among three drone datasets. Our water/vegetation pixels take up a great part, indicating inclusion of rural areas and natural landscapes.

\subsection{Dataset splits and protocol}
In VDD, the images of each scene are split into training, validation and testing sets. VDD contains 280 training, 80 validation and 40 testing images. Our VDD dataset provides three camera angles in each scene. We randomly devide VDD into three subsets, so the subsets should conform to the same data distribution.
All the images of the VDD dataset are made public for researchers to use. We do not hold private test images. Interested researchers may use all 400 images in VDD as the training set or train with other datasets. 

We also make new annotations to UDD\cite{udd} and UAVid\cite{uavid} to fit it with our labeling protocol. The two datasets, combined with ours, compose the Integrated Drone Dataset (IDD). UDD contains 106 images in train set, and 35 in val set. It doesn't have test set. We annotated water in the background class of UDD\cite{udd}. The remaining 6 classes in UDD\cite{udd} is same with VDD, so we don't make other adjustments. In train and val set of UAVid\cite{uavid}, we annotated water based on the background class, and roof based on the building class. Since labels of test set of UAVid\cite{uavid} is not released, we don't include them in IDD.
400 from VDD, 141 from UDD, and 270 from UAVid, IDD contains 811 public available drone images for segmentation. We release the labeled IDD dataset on our website. Class definition of IDD is the same with VDD, so the whole dataset can be trained together. We follow the split of VDD, UDD\cite{udd} and UAVid\cite{uavid}: 536 in train set, 185 in val set, and 40 in test set (only VDD test set). Since the test set size is small, we train models only on train set and validate performance on val set in experiments section. 
Mean Intersection over Union (mIoU) is the recommended evaluation metrics for VDD. We do not officially hold conpetitions, but researchers may honestly use our divided dataset for benchmarking. 

\section{The Integrated IDD Dataset}
As discussed in Section 2, existing drone datasets for semantic segmentations are disparate. None of the individual datasets are sufficiently large. To address this, taking into account domain gap, copyright , and dataset quality, we further annotated the UDD\cite{udd} and UAVid\cite{uavid} datasets using the annotation standards of VDD, resulting in the creation of the Integrated Drone Dataset (IDD).

UDD\cite{udd} and UAVid\cite{uavid} both provides high resolution drone images in urban areas. We label water in UDD\cite{udd} so it is consistent with VDD class definitions. We also label water and roof in UAVid\cite{uavid}. Tree and Low Vegatation are fused into our Vegetation class, Moving Car and Static Car are fused into Vehicles class, Humans class is ignored, so UAVid\cite{uavid} is also adapted into our Integrated Drone Dataset(IDD). All 811 high-resolution images in IDD are taken from drones flying 50 to 120 meters high, and we will prove from experiment that the domain gap in IDD is accpetable in Section 5. We will also show that, learning from IDD will help models understand these three individual datasets better. As has already been discussed in Cityscapes\cite{cityscapes}, extreme weathers shall be tackled with in specific datasets. Therefore, we don't take FloodNet\cite{floodnet} into consideration since it focuses mainly on flooded areas. ICG Drone Dataset\cite{icg}  provides images taken at a height of 30 meters. They mainly focus on individual low-rise houses, so the domain gap is large between ICG Drone Dataset and IDD. There is also copyright issues, so we don't include ICG Drone Dataset in IDD. Images in Aeroscapes\cite{aeroscapes} are only with 720p resolution, and consist of many repetitive scenes. Thus they are not included in IDD. In all, IDD consists of VDD, UDD\cite{udd} and UAVid\cite{uavid} in 7 classes. It is the fruit of 3 public semantic segmentation datasets for low-altitude drones in common scenes, with reasonable gaps within it.

\section{Baselines}
\begin{table*}[h]
	\begin{center}
		\begin{tabular}{|c|cc|cc|cc|}
			\hline
			Dataset & Model & Backbone & Accuracy & mIoU & IDD-Acc & IDD-mIoU \\
			\hline\hline
			VDD & Mask2Former & Swin-T &  89.36 & 77.85 & + 4.64 & + 6.56\\
		    VDD & Mask2Former & ResNet-50 &  94.02 & 83.21 & + 0.41 & + 1.30 \\
            VDD & Segformer-B0 & MiT-B0 &  91.31 & 75.37 & + 1.01 & + 1.14\\
            VDD & Segformer-B2 & MiT-B2 &  92.22 & 85.75 & + 0.59 & + 0.86 \\
            VDD & Segformer-B5 & MiT-B5 &  93.58 & 82.11 & + 0.16 & + 0.26 \\
            VDD & UperNet & Swin-T &  94.20 & 84.73 & + 1.19 & + 0.75 \\
            VDD & UperNet & Swin-L &  92.17 & 85.63 & + 1.86 & + 0.99 \\
            \multicolumn{5}{|l|}{Mean Improvement by training on IDD} & + 1.41 & + 1.69 \\

            \hline\hline
            UDD & Mask2Former & Swin-T &  86.74 & 75.85 & + 2.61 & + 4.50 \\
		    UDD & Mask2Former & ResNet-50 &  86.66 & 75.96 & + 2.43 & + 4.48 \\
            UDD & Segformer-B0 & MiT-B0 &  86.54 & 75.74 & + 0.74 & + 1.29 \\
            UDD & Segformer-B2 & MiT-B2 &  87.01 & 77.49 & + 0.42 & + 0.76 \\
            UDD & Segformer-B5 & MiT-B5 &  88.99 & 79.83 & + 0.13 & + 0.19 \\
            UDD & UperNet & Swin-T &  90.02 & 81.09 & + 0.87 & + 0.40 \\
            UDD & UperNet & Swin-L &  92.15 & 85.61 & + 0.94 & + 0.47 \\
            \multicolumn{5}{|l|}{Mean Improvement by training on IDD} & + 1.16 & + 1.73 \\

            \hline\hline
            UAVid & Mask2Former & Swin-T &  89.11 & 77.7  & + 0.94 & + 1.71 \\
		    UAVid & Mask2Former & ResNet-50 &  89.13 & 77.93 & + 1.14 & + 1.77\\
            UAVid & Segformer-B0 & MiT-B0 &  86.32 & 72.45 & + 0.67 & + 0.95 \\
            UAVid & Segformer-B2 & MiT-B2 &  89.04 & 77.35 & + 0.35 & + 0.71 \\
            UAVid & Segformer-B5 & MiT-B5 &  89.27 & 77.86 & + 0.16 & + 0.22\\
            UAVid & UperNet & Swin-T &  90.35 & 79.61 & + 0.41 & + 0.22\\
            UAVid & UperNet & Swin-L &  90.78 & 80.35 & + 0.53 & + 0.25\\
            \multicolumn{5}{|l|}{Mean Improvement by training on IDD} & + 0.77 & + 0.83 \\

			\hline
		\end{tabular}
	\end{center}
	\caption{Pixel accuracy and mean interaction over union (mIoU) on VDD, UDD\cite{udd} and UAVid\cite{uavid}. Please note that all the datasets used are new versions annotated by us. Each dataset comprises seven categories. Experiments are conducted with state-of-the-art models: Mask2Former\cite{mask2former}, SegFormer\cite{segformer} and UperNet\cite{upernet}. Swin\cite{swin}, ResNet\cite{resnet} and MiT\cite{segformer}are used as backbones. We train the models on train set and test their performance on validation set. IDD-Acc and IDD-mIoU stands for models trained on IDD train set and tested on VDD, UDD\cite{udd} and UAVid\cite{uavid} validation set respectively. We report improvements by training on IDD. All datasets are of 7 classes.}
  	\label{table_experiment}

\end{table*}
In this section, we evaluate the performance of state-of-the-art semantic segmentation models on drone datasets to provide baseline results. We conduct comprehensive experiments on various drone datasets to report the latest results and establish new baselines for future work.
We hope that our experiments will facilitate further research in drone image semantic segmentation.

We train baseline models for 40000 iterations on 4 NVIDIA A10 or 4 GeForce RTX 3090 GPUs, with batch size 3 on each GPU. Segformer B5 \cite{segformer} UperNet \cite{upernet} with Swin-L \cite{swin} are larger models, so we train them with batch size 1 on each GPU for 160000 or more iterations to ensure loss convergence. Such iterations are enough for models to reach an acceptable baseline result, and we note that loss in each experiment converges well, even if train on the largest dataset IDD. 
Our implementation is based on MMSegmentation \cite{mmseg}. Our dataset configuration files in MMSegmentation \cite{mmseg} are released at \href{here}{https://vddvdd.com}.
Basically, we use SGD optimizer with base learning rate 0.01, momentum 0.9 and weight decay 0.0005.
We apply poly learning rate schedulers: $base\_lr * (1-iter/total\_iter) ** 0.9$.
We report final pixel accuracy and  mean intersection over union (mIoU):
\begin{equation}
\text{mIoU} = \frac{1}{N} \sum_{i=1}^{N} \frac{\text{TP}(i)}{\text{FN}(i) + \text{FP}(i) + \text{TP}(i)} .
\end{equation}

\subsection{Experiment on VDD, UDD and UAVid}
State-of-the-art models are trained on VDD, UDD\cite{udd} and UAVid\cite{uavid}. For UDD and UAVid, instead of using the original labels, we use our newly annotated labels with 7 classes in IDD. In other words, each experimented dataset has the same seven classes in Section 3.3 . 

Accuracy and mIoU results in VDD is higher than UDD \cite{udd} and UAVid \cite{uavid}. This is due to difference in dataset collection and splits.
UAVid \cite{uavid} captures 30 videos from 30 different locations and extracts images for annotation. They divide the data into train/val/test sets on a per-video basis. As a result, there is a significant gap between the train/val/test sets in UAVid.
The UDD dataset \cite{udd} was collected from four locations: Peking University, Huludao city, Henan University, and Cangzhou city. Although there is a large gap between the images from different locations, the total number of images is not large (141 images). These images were randomly split into the train and validation sets. We believe that training a model solely on the 106 training images from UDD \cite{udd} may not yield a strong generalization capability.
On the other hand, the proposed VDD dataset randomly divides the data into train/val/test sets, aiming for images to come from the same distribution. The VDD dataset contains a larger number of images, even though they are collected from various locations. As a result, the training of models on VDD yields better results. Additionally, the VDD dataset includes more instances of the "water" and "vegetation" classes (as seen in Fig. \ref{fig_pixel}). We observed that the model achieves higher accuracy on these two classes.
It is important to note that here we are only analyzing the factors affecting the accuracy of the baseline model. The accuracy of a dataset does not necessarily reflect its superiority.

Among the baseline models, Segformer-B0 \cite{segformer} with MiT-B0\cite{segformer} has 3.7 million parameters. Segformer-B2 \cite{segformer} and models with Swin-T \cite{swin} or ResNet50 \cite{resnet} has about 25 to 50 million parameters. Segformer-B5 \cite{segformer} and models with Swin-L \cite{swin} are large models with more than 100 million parameters. The experiments show that on the drone dataset, training larger models does not necessarily yield significantly better results compared to smaller models. Currently, the drone dataset is still relatively small, and the full potential of the larger models has yet to be explored. From small datasets, models may not generalize very well.
This is one of the reasons why we aim for the train/validation/test splits of VDD to follow the same distribution.

\subsection{Experiment on Integrated Drone Dataset}
Integrated Drone Dataset (IDD) provides 536 images in train set. In order to test whether IDD truly helps models to understand visual scenes better, we train the models on IDD train set, and test on VDD, UDD \cite{udd} and UAVid \cite{uavid} validation set. IDD-Acc and IDD-mIoU in Table \ref{table_experiment} reports improvements by training on IDD and testing on individual datasets. 
Training on the IDD dataset for each model consistently yields performance improvements. A larger and more comprehensive dataset in drone image segmentation will encourage more research in this field.

\begin{table*}[h]
	\begin{center}
		\begin{tabular}{|l|l|l|l|}
			\hline
			VDD Class id & VDD Class & UDD \cite{udd} original class & UAVid \cite{uavid} original class  \\
			\hline\hline
			0 & \textbf{other} & other & background clutter\\
			1 & \textbf{wall} & wall & building \\
			2 & \textbf{road} & road & road \\
			3 & \textbf{vegetation} & vegetation & tree and low vegetation \\
			4 & \textbf{vehicle} & vehicle & moving car and static car \\
			5 & \textbf{roof} & roof & \textbf{our new annotation of roof} \\
			6 & \textbf{water} & \textbf{our new annotation of water} & \textbf{our new annotation of water} \\
			\hline
		\end{tabular}
	\end{center}
	
	\caption{Class id comparation table of VDD, UDD \cite{udd} and UAVid \cite{uavid}. Bold text indicates our annotation. We annotate all classes in VDD. For UDD \cite{udd}, we annotate water. For UAVid \cite{uavid}, we annotate roof in building class, and water.}
	\label{table_supp_classid}
\end{table*}

\begin{table*}[h]
	\begin{center}
		\begin{tabular}{|l|ccccccc|c|}
			\hline
			Dataset & other & wall & road & vegetation & vehicle & roof & water & mIoU \\
			\hline\hline
			VDD & 75.76 & 69.01 & 79.07 & 93.11 & 74.25 & 94.27 & 97.02 & 83.21 \\
			UDD \cite{udd} & 89.06 & 68.85 & 68.68 & 63.21 & 85.19 & 60.63 & 96.13 & 75.96 \\
			UAVid \cite{uavid} & 65.64 & 81.72 & 80.89 & 87.6 & 81.1 & 78.02 & 70.50 & 77.93 \\
			\hline
		\end{tabular}
	\end{center}
	\caption{IoU for Mask2Former \cite{mask2former} with ResNet-50 \cite{resnet} backbone on three datasets. We report IoU by training and testing on each dataset.}
	\label{table_supp_iou}
\end{table*}

\begin{table*}[h]
	\begin{center}
		\begin{tabular}{|l|ccccccc|c|}
			\hline
			Dataset & other & wall & road & vegetation & vehicle & roof & water & mIoU \\
			\hline\hline
			VDD & 1.73 & 3.45 & 1.27 & 0.22 & 1.95 & 0.37 & 0.07 & 1.3 \\
			UDD \cite{udd} & 1.79 & 6.93 & 3.06 & 11.56 & 4.05 & 3.87 & 0.01 & 4.48 \\
			UAVid \cite{uavid} & -0.22 & 1.62 & 2.41 & 0.78 & 0.68 & 3.44 & 0.80 & 1.77 \\
			\hline
			Mean Improvement & 1.1 & 8.51 & 2.25 & 4.19 & 2.23 & 2.56 & 0.29 & 2.52 \\
			\hline
		\end{tabular}
	\end{center}
	
	\caption{IoU improvement for Mask2Former \cite{mask2former} with ResNet-50 \cite{resnet} backbone on three datasets. We report IoU improvement by training on IDD, compared to training on each single dataset.}
	\label{table_supp_iou_idd}
\end{table*}

\section{Discussion}
\textbf{Limitation of dataset volume:} The VDD dataset comprises 400 high-resolution images, each with a size of 4000x3000 pixels. While the number of images may appear limited, these aerial scenes are highly complex and contain numerous interconnected objects. Due to such complexity, a direct count of individual objects is not feasible, as objects of the same type often appear connected in the images. Meanwhile, object-level annotation has not been performed in the VDD. For future research, we suggest exploring 3D point cloud segmentation and object-level detection.
From the experiment results of baseline models, it is evident that larger models do not necessarily outperform smaller models on drone datasets. This observation highlights the fact that drone datasets still remain relatively small, restricting the exploration of the full potential of baseline models.
Considering the variance in the dataset, we acknowledge that we have not accounted for the different scenes present in various cities. All the images were captured in temperate cities located in East Asia, thereby missing out on the diversity in building structures across different cultures. This is addressed in IDD, as it includes aerial images from Europe in UAVid \cite{uavid}. Additionally, we have not captured the variance in vegetation that occurs in other climates. It is indeed challenging to encompass such a wide range of scenes within a dataset of only 400 images. Moreover, the personal choices made during drone image capture may have introduced some bias into the VDD.

\textbf{Beyond supervised learning:} Semi-supervised\cite{semi1,semi2,semi3,semi4} and unsupervised\cite{unsup1,unsup2,unsup3,unsup4} semantic segmentation are increasingly popular these days.
These methods, especially unsupervised segmentation algorithms, require a lot of images to train and also suffers from class imbalance in datasets. Considering the small volumn, severe class imbalance, and scene complexity in drone datasets, there are still many challenges to address in performing semi-supervised or unsupervised semantic segmentation on drone images.

\textbf{Broader impacts:} It's important to acknowledge that our paper doesn't explicitly discuss broader impacts in the proposed Varied Drone Dataset (VDD) and Integrated Drone Dataset (IDD), such as fairness or bias. Further research into how our datasets may interact with other aspects of visual scene understanding is encouraged. 

\section{Conclusion}
We propose a novel dataset named Varied Drone Dataset (VDD) for semantic segmentation. VDD comprises 400 densely labeled, high-resolution images, capturing diverse scenes, camera angles, light, and weather conditions. The dataset was collected over the span of a year, encompassing typical objects seen from aerial perspectives, such as residential buildings in urban, industrial, and rural areas, along with facilities like gyms and libraries. Additionally, it includes various natural elements like trees, grasslands, roads, rivers, and lakes. 
Careful consideration was given to select images that minimize overlap while providing a wide range of viewpoints and occlusion relationships, rendering VDD both comprehensive and information-rich.
In order to augment the utility of our dataset, we make new annotations to two existing drone datasets and integrate it with VDD. The Integrated Drone Dataset (IDD) currently stands as the largest drone image segmentation dataset available.

As large-scale datasets play a pivotal role in advancing research, we anticipate that our extensive and diverse dataset will stimulate further interest in enhancing segmentation models and achieving a level of segmentation accuracy for aerial images that is comparable to that of common objects. We believe that this dataset will facilitate the exploration of various vision tasks related to drone images, leveraging precise semantic information.

\section{Comparison table for dataset categories}

We collected and labeled 400 drone images in the Varied Drone Dataset (VDD). Additionally, we combined the existing UDD \cite{udd} and UAVid \cite{uavid} datasets to create the Integrated Drone Dataset (IDD). In the UDD \cite{udd}, we annotated the 'water' class, and in the UAVid \cite{uavid}, we annotated both 'water' and 'roof' classes within the 'building' category. As a result, the VDD, UDD \cite{udd}, and UAVid \cite{uavid} datasets were merged into the IDD dataset, with consistent annotation standards shared with VDD. Table \ref{table_supp_classid} presents the corresponding class mappings among the three datasets.

\section{Class IoU for baseline models}
We present here the IoU (Intersection over Union) for each class within every dataset of Mask2Former \cite{mask2former} with ResNet-50 \cite{resnet} backbone. Table \ref{table_supp_iou} can, to some extent, reflect the difficulty of predictions for each class in the dataset. We also report IoU improvement by training on Integrated Drone Dataset (IDD) in Table \ref{table_supp_iou_idd}. Note that we do not ignore 'other' class (class 0) in the experiments.

\section{Pixel-level labeling with interactive segmentation and Segment Anything Model}
During labeling of dataset, we used LabelMe \cite{labelme} as the labeling tool. We also tried to use interactive segmentation tools \cite{eiseg} and Segment Anything \cite{sam} based tools like  \cite{salt}.  While utilizing annotation tools based on interactive segmentation \cite{eiseg} and SAM \cite{sam}, we encountered unsatisfactory labeling outcomes. Often, even after numerous clicks, the desired regions could not be accurately delineated. As a result, we have opted to continue employing LabelMe \cite{labelme} as our annotation tool.

\section{Dataset usage, metadata and tools}
Please refer to our official website at \url{https://vddvdd.com}. On this website, we have provided the download instructions for the VDD and IDD datasets, information regarding the dataset's licensing terms, basic dataset details, as well as the performance results of various state-of-the-art models on the dataset. We have also shared configuration files based on MMSegmentation~\cite{mmseg}, which can assist users in conducting experiments on our dataset efficiently.

\pagebreak

{\small
	\bibliographystyle{ieee_fullname}
	\bibliography{egbib}
}

\end{document}